\documentclass[pdflatex,sn-mathphys-num]{sn-jnl}

\usepackage{graphicx}%
\usepackage{multirow}%
\usepackage{amsmath,amssymb,amsfonts}%
\usepackage{amsthm}%
\usepackage{mathrsfs}%
\usepackage[title]{appendix}%
\usepackage{xcolor}%
\usepackage{textcomp}%
\usepackage{manyfoot}%
\usepackage{listings}%

\usepackage{float}
\usepackage{multicol} 
\usepackage{multirow}
\usepackage{tabularx}
\usepackage{array}
\usepackage{adjustbox}
\usepackage{cleveref}
\usepackage{indentfirst}
\usepackage{subcaption}
\newcommand{\comment}[1]{} 

\usepackage[linesnumbered,ruled]{algorithm2e}
\usepackage{amsthm} 
\usepackage{makecell} 
\usepackage{setspace}

\usepackage{hhline}
\usepackage{hyperref}
\usepackage{orcidlink}

\usepackage{setspace}

\theoremstyle{thmstyleone}%
\theoremstyle{thmstyletwo}%

\theoremstyle{thmstylethree}%

\begin{document}

\title[Classifier Pooling for Modern Ordinal Classification]{Classifier Pooling for Modern Ordinal Classification}


\author[1]{\fnm{Noam H.} \sur{Rotenberg} \orcidlink{0000-0003-3086-1365}}\email{noam.rotenberg@yale.edu}

\author[2]{\fnm{Andreia V.} \sur{Faria} \orcidlink{0000-0002-1673-002X}}\email{afaria1@jhmi.edu}

\author*[3]{\fnm{Brian} \sur{Caffo} \orcidlink{0000-0002-0793-9497}}\email{bcaffo1@jhu.edu}

\affil[1]{\orgdiv{Department of Biomedical Engineering}, \orgname{Whiting School of Engineering, Johns Hopkins University}, \orgaddress{\city{Baltimore}, \state{MD}, \country{USA}}}

\affil[2]{\orgdiv{Department of Radiology}, \orgname{School of Medicine, Johns Hopkins University}, \orgaddress{\city{Baltimore}, \state{MD}, \country{USA}}}

\affil*[3]{\orgdiv{Department of Biostatistics}, \orgname{Bloomberg School of Public Health, Johns Hopkins University}, \orgaddress{\city{Baltimore}, \state{MD}, \country{USA}}}


\abstract{Ordinal data is widely prevalent in clinical and other domains, yet there is a lack of both modern, machine-learning based methods and publicly available software to address it. In this paper, we present a model-agnostic method of ordinal classification, which can apply any non-ordinal classification method in an ordinal fashion. We also provide an open-source implementation of these algorithms, in the form of a Python package. We apply these models on multiple real-world datasets to show their performance across domains. We show that they often outperform non-ordinal classification methods, especially when the number of datapoints is relatively small or when there are many classes of outcomes. This work, including the developed software, facilitates the use of modern, more powerful machine learning algorithms to handle ordinal data.}

\keywords{ordinal classification, ordinal regression}



\maketitle
\setstretch{1.15}
\section{Introduction}\label{sec1}

Regression and classification are principal approaches to supervised learning. However, certain prediction tasks are not exclusively confined to these approaches; supervised learning tasks where labels are finite and ordered can be framed as ordinal classification (also known as ordinal regression) tasks. Ordinal data is produced in various settings, including staging in pathology and the Likert scale in psychological and consumer surveys.

Ordered logit and ordered probit models are ordinal classification methods introduced in the 1970s \cite{McKelvey-1975_ord, McCullagh-1980_ord_review}. These models are generalizations of linear models, with modifications of the logit and probit link functions, respectively, that extend their application beyond binary data; thresholds on the generalized linear model (to facilitate class assignment) are calculated using maximum likelihood estimation. Implementations are available in Python \cite{seabold2010-statsmodels} and other languages.

The main advantage of these methods is their parsimony and simplicity. They require minimal computing resources because their likelihood optimization is typically convex, and they provide for clear interpretation. However, these methods are often not extended to more powerful, modern machine learning classifiers, such as support vector machine and naïve Bayesian inference. Since computational resources became cheaper and dataset sizes have increased, there is a need for machine learning-based ordinal regression methods with accessible implementation in software.

When these more powerful classifiers are used for ordinal classification tasks, multiclass classification paradigms that ignore ordinality are often used. Classifiers that are inherently binary classifiers (which cannot natively perform multiclass classification) can be adapted for multiclass classification using one vs. rest and one vs. one paradigms. As these methods treat class labels interchangeably, the methods do not take advantage of potentially valuable information encoded in the ranking of classes. On the other hand, non-categorical regression methods are not ideal for ordinal regression tasks because the numerical representation of the rankings are rarely linear; for example, in cancer grading, the difference between stages 1 and 2 should not be considered the same as the difference between stages 2 and 3. Furthermore, regression methods often consider output labels as continuous, which is not approximately true for low numbers of ordinal categories.

Cumulative and hierarchical ordinal classification are paradigms for pooling machine learning classifiers. These paradigms are model-agnostic, which allows for the application of the aforementioned powerful machine learning models that are not natively suited for ordinal classification \cite{tutz2022ordinal_review, Frank2001}. However, to the best of our knowledge, they were not previously implemented in a software library, resulting in their underuse and failure to consider certain practical hyperparameters.

The purpose of this paper is to demonstrate novel approaches to ordinal regression that allow users to select any powerful, modern machine learning classifier that best suits the structure of the dataset. We demonstrate the capabilities of cumulative and hierarchical ordinal regression paradigms and also test novel hyperparameters of these paradigms. To facilitate the use of these methods, an implementation is accessible as a Python package (“statlab”), available on pip (via the command “\texttt{!pip install statlab}”), and is compatible with the sklearn-style classifiers.

\section{Methods}\label{sec2}

\subsection{Model-Agnostic Ordinal Classification Algorithms}
\begin{figure}[hbt!]
\begin{adjustbox}{width=\textwidth,center=\textwidth}
\includegraphics{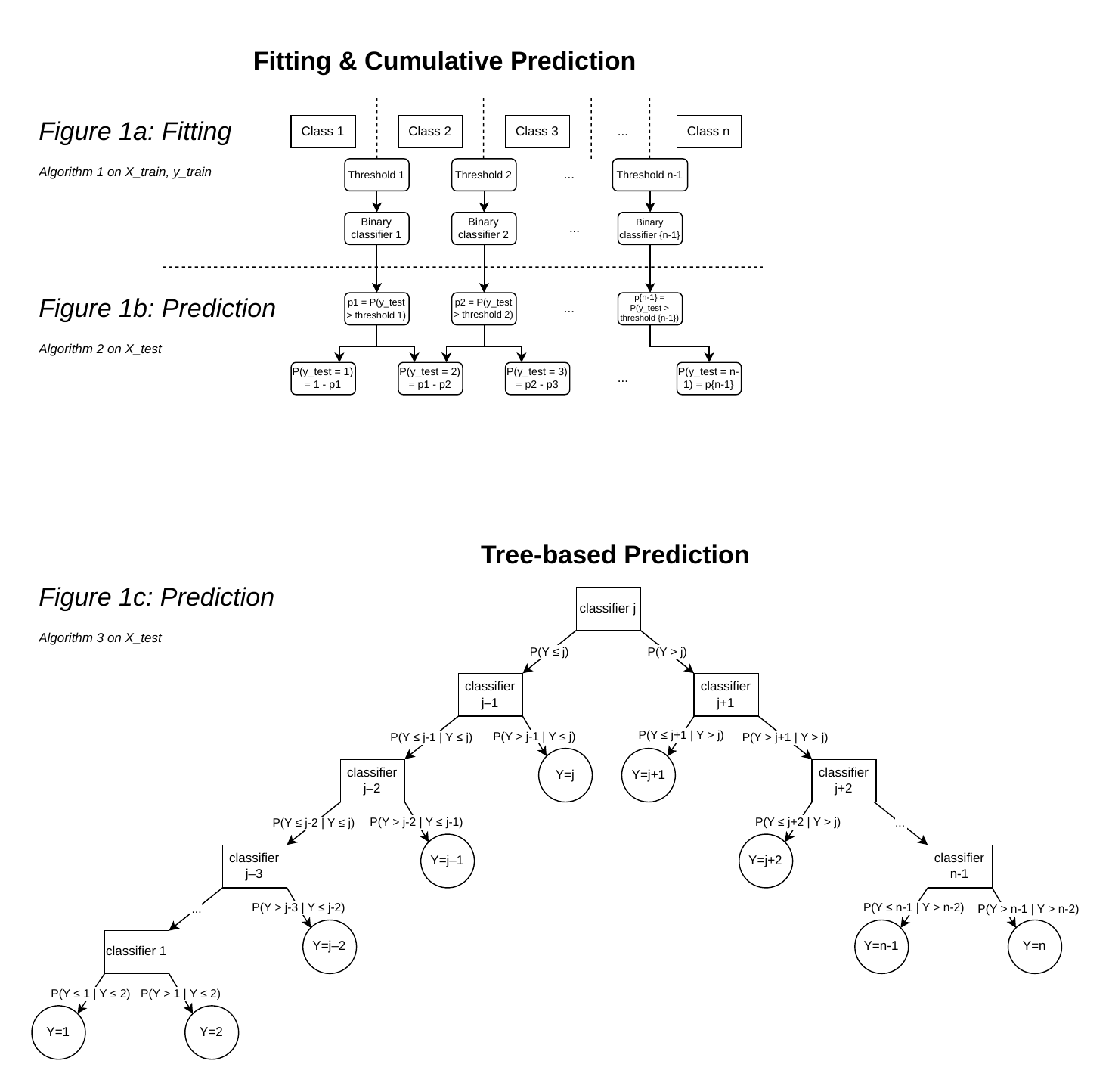}
\end{adjustbox}
\caption{\textit{1a}: Fitting paradigm for thresholded ordinal classification (\textit{Algorithm \ref{alg1:fitting}}), paired with the difference prediction paradigm (fig. \textit{1b}; \textit{Algorithm \ref{alg2:DifferenceOrdinalClassifier}}). \textit{1c}: Prediction paradigm of tree-based ordinal classification (\textit{Algorithm \ref{alg3:TreeOrdinalClassifier}}). To find $P(Y=i)$, multiply all of the conditional probabilities above the node $Y=i$. Equivalence of this product to $P(Y=i)$ is shown in the Appendix.} 
\label{fig:figure1}
\end{figure}

Object-oriented classes were developed in Python to take in any classifier and perform model-agnostic ordinal classification. \texttt{DifferenceOrdinalClassifier()} with default hyperparameters performs cumulative ordinal classification (also called difference-based or subtraction-based ordinal classification), and \texttt{TreeOrdinalClassifier()} with default hyperparameters performs what we refer to as tree-based or hierarchical ordinal classification. Both methods implement \textit{Algorithm~\ref{alg1:fitting}} for learning, where classifiers are each assigned a threshold and trained to determine whether samples are above that threshold, where each threshold corresponds to a pair of adjacent classes. Each classifier is trained on the entirety of the dataset, where each outcome variable is mapped to a binary outcome, signaling whether an input sample is above the classifier's assigned threshold. \texttt{DifferenceOrdinalClassifier()} performs \textit{Algorithm \ref{alg2:DifferenceOrdinalClassifier}} for inference, where the probability of a test sample belonging to a given class is estimated as the difference between the prediction probabilities of the classifiers assigned to the two thresholds adjacent to the given class. \texttt{TreeOrdinalClassifier()} implements \textit{Algorithm \ref{alg3:TreeOrdinalClassifier}} for inference, where the probability of a test sample belonging to a given class is estimated to be iteratively conditioned upon the prediction probabilities of adjacent classes, in a pathway towards the best split index classifier, which is assumed to have the most accurate inference. Graphical representations of \textit{Algorithms \ref{alg1:fitting}}, \textit{\ref{alg2:DifferenceOrdinalClassifier}}, and \textit{\ref{alg3:TreeOrdinalClassifier}} are depicted in \textit{Figures }\hyperref[fig:figure1]{\textit{1a}, \textit{1b}, and \textit{1c}}, respectively.

We developed the Python package \texttt{statlab} to contain these classes and deployed the package on pip. The package performs ordinal classification given any classifier \texttt{clf} for which the functions \texttt{clf.fit(X, y)}, \texttt{clf.predict(X)}, and \texttt{clf.predict\_proba(X)} are compatibly defined. This provides functionality with almost all classifiers from the sklearn library \cite{scikit-learn}. Note that not all sklearn classifiers enable prediction probabilities, so a simpler method is currently implemented in \texttt{statlab}'s \texttt{BaseOrdinalClassifier()}, enabling ordinal classification with truly any base classifier. Note that for simplicity, \textit{Algorithm \ref{alg1:fitting}} displays the “even split” method for calculating the best split index; the default hyperparameter in the software package actually implements the “best classifier” method, which chooses the index with the maximal average F1 score across a 4-fold validation scheme on the training data.

\begin{algorithm}[t]
\caption{\textbf{Fitting classifiers for ordinal classification.} A binary classifier of type \textit{input\_clf} is fit between each adjacent pair of classes. This algorithm presents the “even split” method of choosing the best split index hyperparameter, but various methods for choosing this index are compared in \textit{Experiment~3} below.}
\label{alg1:fitting}

\KwIn{\\
\quad \textit{input\_clf}: binary classification method\\
    \quad $X_{\textit{tr}}$: numeric features table\\
    \quad $y_{\textit{tr}}$: numeric ground-truth labels of $X_{\text{tr}}$\\
\textbf{Requirement:} $(X_{\textit{tr}}, y_{\textit{tr}})$ contains examples for all possible ordinal classes.}
\KwOut{\\
    \quad \textit{classes}: ascending list of unique values in $y_{\textit{tr}}$\\
    \quad \textit{classifiers}: set of classifiers fit to each threshold\\
    \quad \textit{best\_split\_idx}: index of threshold assumed to be the best fit\\
\BlankLine
\textbf{Procedure:}}
\textit{classes} $\gets$ ascending list of unique values in $y_{\textit{tr}}$\;
\textit{thresholds} $\gets$ \textit{classes}$[:-1]$\;
\textit{classifiers} $\gets$ empty list\;
\ForEach{class $c$ \textnormal{in} thresholds}{
    $\textit{clf}_c \gets$ classifier of type \textit{input\_clf}, fit to predict $P(y_{\textit{tr}} > c)$ given $X_{\textit{tr}}$\;
    Append $\textit{clf}_c$ to \textit{classifiers}\;
}
\textit{best\_split\_idx} $\gets \arg\min\limits_i \big| \text{count}(y_{\textit{tr}} > \textit{classes}[i]) - \text{count}(y_{\textit{tr}} \leq \textit{classes}[i]) \big|$\;
\end{algorithm}


\begin{algorithm}[t]
\caption{\textbf{Inference procedure for \texttt{DifferenceOrdinalClassifier()}.} Each classifier is applied to the test sample. A monotonic constraint is applied to the classifier outputs, spreading outward from the threshold indexed by the best split index. The test sample prediction probabilities are computed by subtracting the output of adjacent classifiers.}
\label{alg2:DifferenceOrdinalClassifier}
\KwIn{\\
    \quad \textit{classes}: ascending list of unique values in $y_\textit{tr}$\\
    \quad \textit{classifiers}: set of classifiers fit to each threshold\\
    \quad \textit{best\_split\_idx}: index of threshold assumed to be the best fit\\
    \quad $X_\textit{test}$: test sample features
}
\KwOut{\\
    \quad $\hat{y}_\textit{test\_probs}$: vector of prediction probabilities of the test sample for each class in \textit{classes}\\
    \quad $\hat{y}_\textit{test}$: predicted class of the test sample\\
\BlankLine
\textbf{Procedure:}
}

\textit{thresholds} $\gets$ \textit{classes}$[:-1]$\;
\textit{clf\_probs} $\gets$ empty dictionary\;
\ForEach{\textit{clf}, \textit{thd} \textnormal{pair in} \textit{classifiers} \textnormal{and} \textit{thresholds}}{
    \textit{clf\_probs}[\textit{thd}] $\gets$ estimation of P($y_\textit{test} >$ \textit{thd}) given \textit{clf}\;
}
\ForEach{\textnormal{integer $i$ in the interval $[\textit{best\_split\_idx}, \text{count}(\textit{thresholds}) - 1)$}}{
    $\textit{clf\_probs}[\textit{thresholds}[i+1]] \gets \min(\textit{clf\_probs}[\textit{thresholds}[i]], \textit{clf\_probs}[\textit{thresholds}[i+1]])$\;
}
\ForEach{\textnormal{integer $i$ in the interval $[1, \textit{best\_split\_idx}]$, traversed in reverse}}{
    $\textit{clf\_probs}[\textit{thresholds}[i - 1]] \gets \max(\textit{clf\_probs}[\textit{thresholds}[i - 1]], \textit{clf\_probs}[\textit{thresholds}[i]])$\;
}
$\hat{y}_{\textit{test\_probs}} \gets $ empty list\;
$\hat{y}_\textit{test\_probs}[0] \gets 1 - \textit{clf\_probs}[\textit{thresholds}[0]]$\;
$\hat{y}_\textit{test\_probs}[1:] \gets \text{vector}(\textit{clf\_probs})[1:] - \text{vector}(\textit{clf\_probs})[:-1]$\;
$\hat{y}_\textit{test} \gets \textit{classes}[\arg\max\limits_i(\hat{y}_\textit{test\_probs}[i])]$

\end{algorithm}

\begin{algorithm}[t] 
\caption{\textbf{Inference procedure for \texttt{TreeOrdinalClassifier()}.} Each classifier is applied to the test sample. The test sample prediction probabilities are computed using \textit{Equation~\ref{eq:tree_probs}}, where $y = \textit{best\_split\_idx}$.}
\label{alg3:TreeOrdinalClassifier}
\KwIn{\\
    \quad \textit{classes}: ascending list of unique values in $y_\textit{tr}$\\
    \quad \textit{classifiers}: set of classifiers fit to each threshold\\
    \quad \textit{best\_split\_idx}: index of threshold assumed to be the best fit\\
    \quad $X_\textit{test}$: test sample features
}
\KwOut{\\
    \quad $\hat{y}_\textit{test\_probs}$: vector of prediction probabilities of the test sample for each class in \textit{classes}\\
    \quad $\hat{y}_\textit{test}$: predicted class of the test sample\\
\textbf{Assumptions:}
\begin{enumerate}[noitemsep]
    \item For index $i$ s.t. $i < \textit{best\_split\_idx}$:\\
        \quad \textit{classifiers}[$i$] provides the estimate $P(Y > c_i | Y < c_{i+1})$
    \item For index $i$ s.t. $i > \textit{best\_split\_idx}$: \\ \quad \textit{classifiers}[$i$] provides the estimate $P(Y > c_i | Y > c_{i-1})$

    \end{enumerate}
\\
\BlankLine
\textbf{Procedure:}}

\textit{thresholds} $\gets$ \textit{classes}$[:-1]$\;
\textit{clf\_probs} $\gets$ empty dictionary\;
\ForEach{\textit{clf}, \textit{thd} \textnormal{pair in} \textit{classifiers} \textnormal{and} \textit{thresholds}}{
    $\textit{clf\_probs}[\textit{thd}] \gets$ prediction probability output of \textit{clf} on $X_\textit{test}$\;
}
$\hat{y}_\textit{test\_probs} \gets$ empty list\;
\ForEach{\textnormal{$i$ in the interval $[0, \text{count}(\textit{classes}) )$}}{
    Calculate $\hat{y}_\textit{test\_probs}[i]$ through \textit{Equation 1}, given \textit{clf\_probs}\;
}
$\hat{y}_\textit{test} \gets \textit{classes}[\arg\max\limits_i(\hat{y}_\textit{test\_probs}[i])]$

\end{algorithm}

\subsection{Conditional Equivalence for Tree-based Ordinal Classification}

For an ordinal regression task with the $n$ sorted classes $c_1, c_2, \dots, c_n$, the below equation holds, which is proved in the \hyperref[appendix_A]{\textit{Appendix}}.

\begin{equation}
    P(Y=c_x | y) =
    \begin{cases}
        \left[\prod\limits_{a=1}^{y-1} (1 - P(Y>c_a | Y \le c_{a+1}))\right] \cdot (1 - P(Y > c_y)), & x=1\\
        P(Y > c_{x-1} | Y \le c_x) \cdot \left[\prod\limits_{a=x}^{y-1}  (1 - P(Y > c_a | Y \le c_{a+1})) \right] \cdot (1 - P(Y > c_y)), & 1 < x \le y\\
        P(Y > c_y) \cdot \left[ \prod\limits_{a=y+1}^{x-1} P(Y > c_a | Y > c_{a-1}) \right] \cdot (1 - P(Y > c_k | Y > c_{k-1})), & y < x < n\\
        P(Y > c_y) \cdot \prod\limits_{a=y+1}^{n-1} P(Y > c_a | Y > c_{a-1}), & x=n
    \end{cases}
\label{eq:tree_probs}
\end{equation}

\subsection{Data}
Ordinal regression methods were compared using data acquired from 6 datasets. Dataset 1 contains fetal health data extracted from cardiotocograms of approximately 2,000 patients, which are annotated as normal, suspect pathological, and pathological by three expert obstetricians \cite{cardiotocography_dataset}. Dataset 2 categorizes cars as unacceptable, acceptable, good, or very good while providing the car’s characteristics, such as size, price, and comfort \cite{car_dataset}. Dataset 3 categorizes different wines from 1 to 3 based on quality and provides features, such as hue and concentration of various chemicals \cite{CorCer09-wine}. Dataset 4 is first-order radiomics data extracted from RetinaMNIST using the Python package pyradiomics \cite{pyradiomics}; the original RetinaMNIST is a set of 1600 fundus images of diabetic patient retinas, resized to 3x28x28, and graded 1-5 for diabetic retinopathy severity \cite{medmnistv2}. Dataset 5 is first-order radiomics data extracted from MNIST, which is a set of 28x28 images of hand-written numbers between 0-9 \cite{mnist}; this is used as a negative control, as our method is not expected to improve performance if ordinality is incorporated into a non-ordinal classification task. Dataset 6 is an estimation of the number of rings of an abalone (a type of gastropod), given various body characteristics; the number of rings is correlated with age \cite{abaloneRings_dataset}.

Datasets 1-3 and 6 were randomly split into training and evaluation sets with 70\% and 30\% of the data, respectively. For Datasets 4 and 5, the first 700 training images and first 300 test images were used for training and evaluation, respectively.

\subsection{Evaluation}
In \textit{Experiment 1}, Logistic Regression, Gaussian Naïve Bayes, Support Vector Machine, and Gradient Boosting classifiers were trained on the training sets of Datasets 1-5 through multiple, independent paradigms. The native multiclass methods were used for Gaussian Naïve Bayes and Gradient Boosting classifiers; the one vs. rest (OVR) multiclass paradigm was used for Logistic Regression and Support Vector Classification; the \texttt{sklearn} Python package implementation were used for these models \cite{scikit-learn}. Difference ordinal classification and tree-based ordinal classification were performed for each base classifier, according to Algorithms 1-3 above. Additionally, an ordered logit model and ordered probit model were trained from the Python package \texttt{statsmodels} \cite{seabold2010-statsmodels}. The default hyperparameters from each package were used, to standardize methods. For each model, multiple performance metrics were measured on the evaluation set: accuracy, weighted by inverse class size (AW); polychoric correlation (PC), which is a correlation metric for ordinal data \cite{olsson1979maximum}; and area under the receiver operating characteristic curve, calculated in a one vs. rest paradigm (AUC OVR). This experiment aims to show the general performance of the classification methods across multiple datasets from various domains.

In \textit{Experiment 2}, we tested the performance of various classification methods as the number of ordinal classes and as dataset size varied. Classifiers were independently trained on Dataset 6, where the outcome variable (number of abalone rings) was grouped into either 3 classes ($<$10 rings, between 10-11 rings, and $\ge$12 rings) or 6 classes: [1, 5), [5, 8), [8, 11), [11, 13), [13, 15), [15, 18) and [18, 29] rings; these groupings were formed so that all classes had a substantial number of samples. These two trials were conducted twice: first, with models trained on the entire training set, and second, with models trained on 50\% of the training set of Dataset 6. All trials were evaluated on the same test set. The classifiers trained were the same base models used in \textit{Experiment 1}.

In \textit{Experiment 3}, we evaluated four methods for choosing the best split index, which is an input to \textit{Algorithms \ref{alg2:DifferenceOrdinalClassifier}} and \textit{\ref{alg3:TreeOrdinalClassifier}}. Equivalently, we can choose the threshold that corresponds to this index. One method that has been previously suggested in the literature is arbitrarily choosing the first or last threshold \cite{tutz2022ordinal_review}. Secondly, the most center threshold could be chosen. Another method, which is shown in \textit{Algorithm \ref{alg1:fitting}}, is to select the threshold where data is most evenly balanced. Additionally, the threshold with the best classification performance on a held-out validation set could be selected; this could hypothetically minimize incorrect adjustments that are applied to the other models’ outputs in \textit{Algorithm \ref{alg2:DifferenceOrdinalClassifier}} and provide the most robust probability estimation to be conditioned upon by the other models’ probabilities in \textit{Algorithm \ref{alg3:TreeOrdinalClassifier}}. This fourth option is implemented in a 4-fold validation paradigm during the classifier fitting phase. The four methods were evaluated on Dataset 6, using the base models from \textit{Experiment 1} in difference ordinal classification and tree-based ordinal classification paradigms. The outcome variables (number of abalone rings) were regrouped into: [1, 5), [5, 7), [7, 8), [8, 9), [9, 10), [10, 12), [12, 15) and [15, 29]. The purpose of this manipulation was to distribute the outcome classes non-uniformly, so that the “even split” and “middle index” methods resulted in different indices. A summary of the experiments can be found in \textit{Table \ref{tab:table1-experiments}}.

\begin{table}[]
\begin{adjustbox}{width=1\textwidth}
\begin{tabular}{|l|lll|}
\hline
\textbf{}                  & \multicolumn{1}{l|}{\textbf{Experiment 1}}                                                                                                             & \multicolumn{1}{l|}{\textbf{Experiment 2}}                                                                                                                       & \textbf{Experiment 3}                                                                                                                                                                                                                                                                                                                                                                                                  \\ \hline
\textbf{Target comparison} & \multicolumn{1}{l|}{\makecell[l]{Performance of ordinal\\ vs. non-ordinal classifiers\\ across   various domains}}                                                    & \multicolumn{1}{l|}{\makecell[l]{Performance of ordinal\\ vs. non-ordinal classifiers as\\ dataset   size changes and\\ number of classes changes}}                                  & \makecell[l]{Performance of difference and\\ tree-based ordinal   classification\\ as the best split index\\ hyperparameter changes}                                                                                                                                                                                                                                                                                                      \\ \hline
\textbf{Datasets}          & \multicolumn{1}{l|}{\makecell[l]{Datasets 1-5: fetal health,\\ car quality, wine quality,\\   RetinaMNIST radiomics,\\ MNIST radiomics}}                                   & \multicolumn{1}{l|}{Dataset 6: abalone rings}                                                                                                                    & Dataset 6: abalone rings                                                                                                                                                                                                                                                                                                                                                                                      \\ \hline
\textbf{Classifier types}  & \multicolumn{2}{l|}{\begin{tabular}[l]{@{}l@{}}Native multiclass and ordinal classification methods for:\\ 	\ • Logistic Regression\\ \ • 	Gaussian Naïve Bayes\\ \ • 	Support Vector Classification\\ \ • 	Gradient Boosting Classifier\\ Additionally:\\ \ • 	Ordered logit model\\ \ • 	Ordered probit model\end{tabular}} 
& \begin{tabular}[l]{@{}l@{}}Difference and tree-based ordinal\\ classification \textit{only},   for:\\     \ •  Logistic  Regression\\    \ •  Gaussian Naïve Bayes\\    \ •  Support Vector Classification\\    \ •  Gradient Boosting Classifier\end{tabular} \\ \hline
\textbf{Metrics}           & \multicolumn{3}{l|}{\makecell[l]{Weighted accuracy;\\ polychoric correlation;\\ area under the   receiving operating characteristic curve, one vs. rest}}                                                                                                                                                                                                                                                                                                                                                                                                                                                                                                                                                                                                   \\ \hline
\end{tabular}
    
\end{adjustbox}
\caption{Summary of \textit{Experiments 1} through \textit{3}.}
\label{tab:table1-experiments}
\end{table}

\section{Results}

The raw classifier metrics for \textit{Experiments 1}, \textit{2}, and \textit{3} are reported in \textit{Tables \ref{tab:table2-exp1}}, \textit{\ref{tab:table3-exp2_raw}}, and \textit{\ref{tab:table5-exp3}}, respectively. Even without hyperparameter tuning, our methods exhibit improved performance in the majority of metrics throughout all tested datasets. In \textit{Experiment 1}, difference and tree-based ordinal classification models met or outperformed the native base classifier in 36/48 (75\%) and 38/48 (79\%) of metrics, respectively, across the ordinal datasets 1 through 4. They also frequently outperformed the ordered logit and probit models. Notably, the tree-based ordinal classification method met or outperformed the traditional models in all of the metrics in Dataset 2 but performed approximately as well as them in dataset 3, reinforcing the notion that model performance is highly dataset-dependent. Notably, some traditional models failed to achieve better-than-chance performance on the RetinaMNIST radiomics dataset, while none of the difference or tree-based ordinal classification methods failed to train on the ordinal datasets (1-4).

In Dataset 5, the MNIST radiomics “negative control,” two of the ordinal models failed to achieve better-than-chance performance, and the other ordinal regression methods frequently performed worse than the traditional classification methods. This suggests that providing ordinal tags to non-ordinal classes decreases classification performance; while MNIST ground-truth labels are digits 0 through 9, they would more appropriately be referred to as strings than numbers.

\begin{table}
\centering
\resizebox{\linewidth}{!}{%
\begin{tabular}{|ll|ccc|ccc|ccc|ccc|ccc|}
\hline
                       &                     & \multicolumn{3}{c|}{\makecell{Dataset
  1:\\ Fetal Health}} & \multicolumn{3}{c|}{\makecell{Dataset 2:\\ Car Quality}} & \multicolumn{3}{c|}{\makecell{Dataset 3:\\ Wine
  Quality}} & \multicolumn{3}{c|}{\makecell{Dataset 4:\\ RetinaMNIST radiomics}} & \multicolumn{3}{c|}{\makecell{Dataset 5:\\ MNIST
  radiomics\\ (negative control)}}  \\
Method                 & Method Subtype      & AW    & PC    & \makecell{AUC\\ OVR}                        & AW    & PC    & \makecell{AUC\\ OVR}                     & AW    & PC    & \makecell{AUC\\ OVR}                        & AW    & PC    & \makecell{AUC\\ OVR}                               & AW    & PC    & \makecell{AUC\\ OVR}                                              \\ 
\hline
Logistic Regression    & OVR                 & 0.744 & 0.914 & 0.925                          & 0.657 & 0.899 & 0.963                       & 0.968 & 0.999 & 0.998                          & 0.215 & 0.367 & 0.637                                 & 0.224 & 0.280 & 0.690                                                \\
                       & Ordered logit model & 0.698 & 0.911 & 0.947                          & 0.607 & 0.909 & 0.945                       & 0.908 & 0.999 & 0.929                          & \multicolumn{3}{c|}{*}                                & \multicolumn{3}{c|}{*}                                                \\
                       & Difference          & 0.757 & 0.906 & 0.936                          & 0.686 & 0.907 & 0.966                       & 0.984 & 0.999 & 0.998                          & 0.306 & 0.594 & 0.698                                 & 0.221 & 0.294 & 0.690                                                \\
                       & Tree-based          & 0.763 & 0.912 & 0.935                          & 0.662 & 0.913 & 0.964                       & 0.984 & 0.999 & 0.997                          & 0.320 & 0.604 & 0.708                                 & 0.191 & 0.325 & 0.648                                                \\ 
\hline
\multicolumn{2}{|l|}{Ordered probit model}    & 0.702 & 0.914 & 0.947                          & 0.586 & 0.910 & 0.945                       & 0.908 & 0.999 & 0.930                          & \multicolumn{3}{c|}{*}                                & 0.124 & 0.126 & 0.528                                                \\ 
\hline
Gaussian
  Naïve Bayes & Multiclass          & 0.762 & 0.831 & 0.900                          & 0.622 & 0.894 & 0.932                       & 0.966 & 0.999 & 1.000                          & 0.265 & 0.383 & 0.640                                 & 0.192 & 0.238 & 0.640                                                \\
                       & Difference          & 0.775 & 0.879 & 0.933                          & 0.754 & 0.959 & 0.871                       & 0.966 & 0.999 & 0.998                          & 0.248 & 0.509 & 0.605                                 & 0.192 & 0.120 & 0.635                                                \\
                       & Tree-based          & 0.775 & 0.879 & 0.934                          & 0.708 & 0.942 & 0.962                       & 0.966 & 0.999 & 0.998                          & 0.285 & 0.425 & 0.653                                 & 0.178 & 0.127 & 0.611                                                \\ 
\hline
Support Vector         & OVR                 & 0.612 & 0.845 & 0.931                          & 0.812 & 0.975 & 0.995                       & 0.605 & 0.446 & 0.866                          & \multicolumn{3}{c|}{*}                                & 0.200 & 0.202 & 0.649                                                \\
Classification             & Difference          & 0.716 & 0.874 & 0.931                          & 0.930 & 0.983 & 0.993                       & 0.668 & 0.637 & 0.892                          & 0.292 & 0.557 & 0.644                                 & 0.111 & 0.219 & 0.580                                                \\
                       & Tree-based          & 0.692 & 0.845 & 0.930                          & 0.909 & 0.982 & 0.995                       & 0.598 & 0.731 & 0.879                          & 0.299 & 0.536 & 0.666                                 & \multicolumn{3}{c|}{*}                                                \\ 
\hline
Gradient
  Boosting    & Multiclass          & 0.886 & 0.964 & 0.984                          & 0.948 & 0.986 & 0.998                       & 0.978 & 0.999 & 1.000                          & 0.316 & 0.551 & 0.638                                 & 0.205 & 0.207 & 0.682                                                \\
Classification             & Difference          & 0.903 & 0.973 & 0.978                          & 0.955 & 0.986 & 0.999                       & 0.959 & 0.999 & 0.976                          & 0.311 & 0.531 & 0.623                                 & 0.217 & 0.266 & 0.652                                                \\
                       & Tree-based          & 0.903 & 0.973 & 0.981                          & 0.955 & 0.986 & 0.998                       & 0.959 & 0.999 & 0.986                          & 0.303 & 0.482 & 0.662                                 & 0.234 & 0.308 & 0.659                                               \\ \hline
\end{tabular}
}
\caption{Results of \textit{Experiment 1}; evaluation of classification methods, including difference and tree-based ordinal classification, across various datasets with ordinal outcome variables. Models that failed to train (defined as when the AW was less than chance) were marked by *. AW = accuracy, weighted by inverse class size; PC = polychoric correlation; OVR = one vs. rest; AUC OVR = multi-class area under the receiver operating characteristic curve, calculated with an OVR paradigm.}
\label{tab:table2-exp1}
\end{table}

In \textit{Experiment 2}, many performance metrics decreased as the training set size decreased or as the number of outcome classes increased (\textit{Table \ref{tab:table3-exp2_raw}}). However, the decrease in performance of the difference and tree-based ordinal classification methods was generally smaller than that of the other methods. \textit{Table \ref{tab:table4-exp2_difference}} shows the change in performance from the full train set, 3-class groupings task to the 50\%-reduced train set, 7-class groupings task; the decrease in performance was smaller (i.e., less bad) for difference and tree-based ordinal regression in comparison to the native base classifier for 8/12 (67\%) and 8/12 (67\%) of the metrics, respectively. Note that polychoric correlation increased as the number of outcome classes increased; this may be due to enabling the model to provide a higher level of granularity with respect to the difference between different samples, even if the exact categorization predicted by the model is incorrect.

\begin{table}
\begin{adjustbox}{width=1\textwidth}
\centering
\resizebox{\linewidth}{!}{%
\begin{tabular}{|lll|ccc|ccc|}
\hline
 &  &  & \multicolumn{3}{c|}{Full training set} & \multicolumn{3}{c|}{50\% of training set} \\
\makecell{Training\\ Paradigm} & Method & Method Subtype & AW & PC & \makecell{AUC\\ OVR} & AW & PC & \makecell{AUC\\ OVR} \\ 
\hline
3 outcome & Logistic  Regression & OVR & 0.594 & 0.721 & 0.812 & 0.590 & 0.712 & 0.803 \\
classes &  & Ordered logit model & 0.571 & 0.724 & 0.798 & 0.576 & 0.737 & 0.797 \\
 &  & Difference & 0.583 & 0.711 & 0.806 & 0.561 & 0.667 & 0.796 \\
 &  & Tree-based & 0.561 & 0.692 & 0.802 & 0.533 & 0.669 & 0.790 \\ 
\hhline{~--------}
 & Ordered probit model &  & 0.568 & 0.727 & 0.797 & 0.566 & 0.735 & 0.797 \\ 
\hhline{~--------}
 & Gaussian Naïve Bayes & Multiclass & 0.506 & 0.578 & 0.739 & 0.495 & 0.565 & 0.740 \\
 &  & Difference & 0.483 & 0.559 & 0.710 & 0.482 & 0.560 & 0.710 \\
 &  & Tree-based & 0.480 & 0.559 & 0.706 & 0.486 & 0.562 & 0.708 \\ 
\hhline{~--------}
 & Support Vector Classification & OVR & 0.565 & 0.680 & 0.816 & 0.503 & 0.614 & 0.799 \\
 &  & Difference & 0.592 & 0.694 & 0.809 & 0.572 & 0.657 & 0.797 \\
 &  & Tree-based & 0.579 & 0.689 & 0.811 & 0.565 & 0.653 & 0.798 \\ 
\hhline{~--------}
 & Gradient Boosting Classification & Multiclass & 0.619 & 0.750 & 0.828 & 0.607 & 0.730 & 0.817 \\
 &  & Difference & 0.616 & 0.751 & 0.825 & 0.603 & 0.732 & 0.815 \\
 &  & Tree-based & 0.615 & 0.761 & 0.829 & 0.604 & 0.738 & 0.820 \\ 
\hline
7 outcome & Logistic Regression & OVR & 0.241 & 0.704 & 0.845 & 0.236 & 0.695 & 0.837 \\
classes &  & Ordered logit model & 0.296 & 0.763 & 0.833 & 0.299 & 0.749 & 0.831 \\
 &  & Difference & 0.259 & 0.715 & 0.844 & 0.261 & 0.701 & 0.833 \\
 &  & Tree-based & 0.269 & 0.723 & 0.841 & 0.260 & 0.707 & 0.829 \\ 
\hhline{~--------}
 & \multicolumn{2}{l|}{Ordered probit model} & 0.259 & 0.750 & 0.831 & 0.256 & 0.740 & 0.829 \\ 
\hhline{~--------}
 & Gaussian Naïve Bayes & Multiclass & 0.404 & 0.646 & 0.764 & 0.424 & 0.653 & 0.768 \\
 &  & Difference & 0.394 & 0.654 & 0.722 & 0.398 & 0.656 & 0.725 \\
 &  & Tree-based & 0.383 & 0.646 & 0.734 & 0.383 & 0.648 & 0.735 \\ 
\hhline{~--------}
 & Support Vector & OVR & 0.242 & 0.742 & 0.857 & 0.236 & 0.717 & 0.855 \\
 & Classification & Difference & 0.351 & 0.764 & 0.826 & 0.367 & 0.733 & 0.817 \\
 &  & Tree-based & 0.344 & 0.745 & 0.857 & 0.350 & 0.721 & 0.855 \\ 
\hhline{~--------}
 & Gradient Boosting & Multiclass & 0.347 & 0.748 & 0.837 & 0.335 & 0.731 & 0.818 \\
 & Classification & Difference & 0.356 & 0.768 & 0.825 & 0.362 & 0.747 & 0.795 \\
 &  & Tree-based & 0.353 & 0.768 & 0.855 & 0.352 & 0.739 & 0.842\\
 \hline
\end{tabular}
}
\end{adjustbox}
\caption{Results of \textit{Experiment 2}; comparison of classification methods, including difference and tree-based ordinal classification, on Dataset 6 (abalone rings), while varying dataset size and number of classes. AW = accuracy, weighted by inverse class size; PC = polychoric correlation; OVR = one vs. rest; AUC OVR = multi-class area under the receiver operating characteristic curve, calculated with an OVR paradigm.}
\label{tab:table3-exp2_raw}
\end{table}

\begin{table}
\centering
\begin{tabular}{|llccc|} 
\hline
Method & Method Subtype & $\Delta$AW & $\Delta$PC & $\Delta$AUC OVR \\ 
\hline
Logistic Regression & OVR & -0.358 & -0.026 & ~0.025 \\
 & Ordered logit model & -0.272 & ~0.025 & ~0.033 \\
 & Difference & -0.322 & -0.010 & ~0.027 \\
 & Tree-based & -0.302 & ~0.015 & ~0.028 \\ 
\hline
\multicolumn{2}{|l}{Ordered probit model} & -0.311 & ~0.012 & ~0.031 \\ 
\hline
Gaussian Naïve Bayes & Multiclass & -0.082 & ~0.074 & ~0.029 \\
 & Difference & -0.084 & ~0.098 & ~0.015 \\
 & Tree-based & -0.097 & ~0.089 & ~0.028 \\ 
\hline
Support Vector Classification & OVR & -0.329 & ~0.037 & ~0.040 \\
 & Difference & -0.226 & ~0.039 & ~0.008 \\
 & Tree-based & -0.229 & ~0.032 & ~0.044 \\ 
\hline
Gradient
  Boosting Classification & Multiclass & -0.284 & -0.018 & -0.010 \\
 & Difference & -0.254 & -0.004 & -0.029 \\
 & Tree-based & -0.264 & -0.022 & ~0.013 \\
\hline
\end{tabular}
\caption{Results of \textit{Experiment 2}; change in performance from 3 outcome classes on the full dataset to 7 outcome classes on the 50\% dataset (i.e., performance on 7-class groupings on 50\% of the train dataset minus performance on 3-class groupings on full train dataset). $\Delta$AW = change in accuracy, weighted by inverse class size; $\Delta$PC = change in polychoric correlation; OVR = one vs. rest; $\Delta$AUC OVR = change in multi-class area under the receiver operating characteristic curve, calculated with an OVR paradigm.}
\label{tab:table4-exp2_difference}
\end{table}

In \textit{Experiment 3}, using the middle index as the best split index hyperparameter outperformed the other indices, achieving the maximum classifier performance in 7/24 cases. The other methods followed closely behind each other, with the best classifier, last index, and even split methods achieving the maximum metric in 3/24, 3/24, and 1/24 cases, respectively. There were multiple instances (7/24) in which all classifiers achieved the same performance.

\begin{table}
\centering
\resizebox{\linewidth}{!}{%
\begin{tabular}{|ll|ccc|ccc|ccc|ccc|} 
\hline
\makecell{Best split index\\ calculation method} &  & \multicolumn{3}{c|}{Even Split} & \multicolumn{3}{c|}{\makecell{Best
  Classifier\\ (via 4-fold validation)**}} & \multicolumn{3}{c|}{Last index} & \multicolumn{3}{c|}{Middle index} \\
 & Best split index & \multicolumn{3}{c|}{4} & \multicolumn{3}{c|}{0***} & \multicolumn{3}{c|}{6} & \multicolumn{3}{c|}{3} \\
Method & Method Subtype & AW & PC & \makecell{AUC\\ OVR} & AW & PC & \makecell{AUC\\ OVR} & AW & PC & \makecell{AUC\\ OVR} & AW & PC & \makecell{AUC\\ OVR} \\ 
\hline
Logistic Regression & Difference & 0.310 & 0.767 & 0.824 & 0.310 & 0.767 & 0.824 & 0.310 & 0.767 & 0.824 & 0.310 & 0.767 & 0.824 \\
 & Tree-based & 0.268 & 0.764 & 0.813 & \underline{0.307} & 0.746 & 0.815 & 0.269 & \underline{0.782} & \underline{0.816} & 0.285 & 0.747 & 0.814 \\ 
\hline
Gaussian Naïve Bayes & Difference & 0.346 & 0.689 & 0.731 & 0.346 & 0.689 & 0.731 & 0.346 & 0.689 & 0.731 & 0.346 & 0.689 & 0.731 \\
 & Tree-based & 0.342 & 0.684 & 0.740 & 0.342 & 0.685 & 0.736 & 0.342 & 0.683 & 0.734 & 0.342 & 0.685 & \underline{0.741} \\ 
\hline
Support Vector & Difference & 0.359 & 0.762 & 0.808 & 0.363 & 0.763 & 0.809 & 0.357 & 0.759 & 0.805 & \underline{0.371} & \underline{0.764} & \underline{0.810} \\
Classification & Tree-based & 0.336 & 0.757 & \underline{0.820} & \underline{0.353} & 0.760 & 0.818 & 0.344 & \underline{0.774} & 0.813 & 0.339 & 0.759 & 0.819 \\ 
\hline
Gradient Boosting & Difference & 0.376 & 0.772 & 0.820 & 0.381 & 0.770 & 0.819 & 0.378 & 0.762 & 0.817 & 0.381 & \underline{0.774} & 0.820 \\
Classification & Tree-based & 0.366 & 0.794 & 0.833 & \underline{0.380} & 0.783 & 0.830 & 0.369 & 0.781 & 0.833 & 0.373 & \underline{0.795} & \underline{0.834} \\
\hline
\end{tabular}
}
\caption{Results of \textit{Experiment 3}; comparison of methods for choosing the best split index hyperparamter on Dataset 6 (abalone rings). The underlined values are the maximum for each metric, for each classifier, for each method, if the maximum value is unique. AW = accuracy, weighted by inverse class size; PC = polychoric correlation; OVR = one vs. rest; AUC OVR = multi-class area under the receiver operating characteristic curve, calculated in an OVR paradigm.\\
**Note that the results for this hyperparameter are not the same as in \textit{Experiment 2} because the variables were regrouped, as described in \textit{Methods: Evaluation}.\\
***Note that the index is dependent on classifier performance, but in this case, the index was 0 for all classifiers.}
\label{tab:table5-exp3}
\end{table}

\section{Discussion}

In this paper, we evaluate methods for ordinal classification on real-world datasets, including two clinical datasets. The improved performance of ordinal classification over non-ordinal classification is marginal for some datasets and large for others, but overall shows that there is substantial information encoded in the ordinality of the classes that a non-ordinal multiclass paradigm cannot consider. Out of all metrics, ordinal classification usually had greater polychoric correlation, as non-ordinal classification does not have any mechanism to optimize this metric. The tree-based ordinal classification method performed marginally better than the cumulative ordinal classification method, so it might more robustly represent classifiers’ estimated probabilities. Surprisingly, the choice of the best split index, might be somewhat arbitrary given the results of \textit{Experiment 3}. The “middle index” had stronger performance over the other indices, but additional investigation is necessary to determine whether this trend holds true in other datasets. In particular, more investigation of the “best classifier” approach would be needed before the additional computation time could be deemed justified. Overall, these novel approaches may be useful when considering the practical limitations of real-world data, as some thresholds might have fewer datapoints than others or may have poorer performance for other reasons.

In addition, we provide a novel software implementation, in the Python package \texttt{statlab}. Our implementation is model-agnostic; it is compatible with most sklearn classifiers, and any other classifier with three functions, of the form: \texttt{clf.fit(X, y)}, \texttt{clf.predict(X)}, and \texttt{clf.predict\_proba(X)}. The deployment of open-source software and the provided interoperability allows for easy adoption, replication, and testing of methods, fulfilling the principles of open science.

A limitation of this work is that it is less useful for deep learning. Neural networks have much more complicated decision rules that may be inherently adaptable to ordinal classification tasks. However, these ordinal classification methods are still very useful for cases when data is sparse—as emphasized by \textit{Experiment 2}—and may still allow for using shallow multi-layer perceptrons.

An additional limitation is that model performance is highly dependent on the dataset. Shown in \textit{Experiment 1}, ordinal models performed better on some datasets than others, when compared with the non-ordinal models, despite all of the datasets having ordinal outcomes (except for MNIST, the negative control). However, this limitation is inherent to machine learning, and machine learning researchers should always evaluate many methodologies to develop a system that best fits their data. In our implementation, we provide choices of difference or tree-based ordinal classification, unlimited choices of base models, and multiple choices for additional hyperparameters. Overall, this work---including comparisons, software implementation, and hyperparameter development---enables simple, intuitive, but powerful modification of classification schemes to improve model performance in ordinal regression tasks.

\section*{Data and Code Availability}
All data used is publicly accessible. See sources: \cite{cardiotocography_dataset, car_dataset, CorCer09-wine, medmnistv2, mnist, abaloneRings_dataset}.

All code used to produce results reported in this manuscript are available in Zenodo: \url{https://doi.org/10.5281/zenodo.18990527} \cite{rotenberg_2026_18990527}.

Updated versions of the software may be found in the \texttt{statlab} github and pypi package: \url{https://github.com/noamrotenberg/statlab}, \url{https://pypi.org/project/statlab}.

\section*{Acknowledgements}
This research was supported in part by the National Institute of Deaf and Communication Disorders, NIDCD, through R01 DC05375, R01 DC015466, P50 DC014664, the National Institute of Biomedical Imaging and Bioengineering, NIBIB, through P41 EB031771.

\section*{Competing interests}
The authors report no competing interests.

\bibliography{reference}

\appendix
\section*{Appendix: Proof of Equation 1} \label{appendix_A}

For an ordinal regression task with the $n$ sorted classes $c_1, c_2, \dots, c_n$, \textit{Equation \ref{eq:tree_probs}} asserts:

$$
    P(Y=c_x | y) =
    \begin{cases}
        \left[\prod\limits_{a=1}^{y-1} (1 - P(Y>c_a | Y \le c_{a+1}))\right] \cdot (1 - P(Y > c_y)), & x=1\\
        P(Y > c_{x-1} | Y \le c_x) \cdot \left[\prod\limits_{a=x}^{y-1}  (1 - P(Y > c_a | Y \le c_{a+1})) \right] \cdot (1 - P(Y > c_y)), & 1 < x \le y\\
        P(Y > c_y) \cdot \left[ \prod\limits_{a=y+1}^{x-1} P(Y > c_a | Y > c_{a-1}) \right] \cdot (1 - P(Y > c_k | Y > c_{k-1})), & y < x < n\\
        P(Y > c_y) \cdot \prod\limits_{a=y+1}^{n-1} P(Y > c_a | Y > c_{a-1}), & x=n
    \end{cases}
$$

\textit{Proof.} Given sorted ordinal classes $c_1$, $c_2$, \dots $c_i$, \dots $c_j$, \dots $c_k$, \dots, $c_n$:
Let $Y$ be the class assignment of a sample of interest, which is a random variable.

Expanding $P(Y = c_k)$:\\
$P(Y > c_{j+1}) = P(Y > c_{j+1}, Y > c_j) = P(Y > c_{j+1} | Y > c_j) \cdot P(Y > c_j)\\
P(Y = c_{j+1}) = P(Y \le c_{j+1}, Y > c_j) = (1 - P(Y > c_{j+1} | Y > c_j)) \cdot P(Y > c_j)\\
P(Y > c_{j+2}) = P(Y > c_{j+2}, Y > c_{j+1}) = P(Y > c_{j+2} | Y > c_{j+1}) \cdot P(Y > c_{j+1}) = P(Y > c_{j+2} | Y > c_{j+1}) \cdot P(Y > c_{j+1} | Y > c_j) \cdot P(Y > c_j)$\\
Similarly, $P(Y > c_k) = P(Y > c_k, Y > c_{k-1}) = P(Y > c_k | Y > c_{k-1}) \cdot P(Y > c_{k-1}) = \prod\limits_{a=j+1}^k [P(Y > c_a | Y > c_{a-1})] \cdot P(Y > c_j)$\\
Consequently, $P(Y = c_k) = P(Y > c_{k-1}, Y \le c_k) = P(Y \le c_k | Y > c_{k-1}) \cdot P(Y > c_{k-1}) = P(Y > c_j) \cdot \prod\limits_{a=j+1}^{k-1} [P(Y > c_a | Y > c_{a-1})] \cdot (1 - P(Y > c_k | Y > c_{k-1}))$

Expanding $P(Y = c_n)$:\\
$P(Y = c_n) = P(Y > c_{n-1}) = P(Y > c_j) \cdot \prod\limits_{a=j+1}^{n-1} [P(Y > c_a | Y > c_{a-1})]$

Expanding $P(Y = c_i)$:\\
$P(Y \le c_{j-1}) = P(Y \le c_{j-1}, Y \le c_j) = P(Y \le c_{j-1} | Y \le c_j) \cdot P(Y \le c_j) = (1 - P(Y > c_{j-1} | Y \le c_j)) \cdot (1 - P(Y > c_j))\\
P(Y \le c_{j-2}) = P(Y \le c_{j-2}, Y \le c_{j-1}) = P(Y \le c_{j-2} | Y \le c_{j-1}) \cdot P(Y \le c_{j-1}) = (1 - P(Y > c_{j-2} | Y \le c_{j-1})) \cdot (1 - P(Y > c_{j-1} | Y \le c_j)) \cdot (1 - P(Y > c_j))$\\
Similarly, $P(Y \le c_i) = \prod\limits_{a=i}^{j-1} [(1 - P(Y > c_a | Y \le c_{a+1})] \cdot (1 - P(Y > c_j))$\\
Consequently, $P(Y = c_i) = P(Y \le c_i, Y > c_{i-1}) = P(Y > c_{i-1} | Y \le c_i) \cdot P(Y \le c_i) = P(Y > c_{i-1} | Y \le c_i) \cdot \prod\limits_{a=i}^{j-1} [(1 - P(Y > c_a | Y \le c_{a+1})] \cdot (1 - P(Y > c_j))$

Expanding $P(Y = c_1)$:\\
$P(Y = c_1) = P(Y \le c_1) = \prod\limits_{a=1}^{j-1} [(1 - P(Y > c_a | Y \le c_{a+1})] \cdot (1 - P(Y > c_j))$ \ \qedsymbol

\end{document}